\documentclass[10pt,twocolumn,letterpaper]{article}

\usepackage[pagenumbers]{cvpr} 

\usepackage{graphicx}
\usepackage{amsmath}
\usepackage{amssymb}
\usepackage{booktabs}
\usepackage{makecell}
\usepackage{array,hhline}

\usepackage{comment}
\usepackage{color}
\usepackage{overpic}
\usepackage{multirow}
\usepackage{tabularx}
\usepackage{adjustbox}
\usepackage{times}
\usepackage{epsfig}
\usepackage{float}

\usepackage[pagebackref,breaklinks,colorlinks]{hyperref}

\usepackage[capitalize]{cleveref}
\crefname{section}{Sec.}{Secs.}
\Crefname{section}{Section}{Sections}
\Crefname{table}{Table}{Tables}
\crefname{table}{Tab.}{Tabs.}

\def\cvprPaperID{1778} 
\def\confName{CVPR}
\def\confYear{2023}

\begin{document}

\title{Spatial-Frequency Attention for Image Denoising}

\author{Shi Guo\thanks{Equal contribution} \quad Hongwei Yong\textsuperscript{*} \quad Xindong Zhang \quad Jianqi Ma \quad Lei Zhang \\
The Hong Kong Polytechnic University \\
{\tt\small shiguo.guo@connect.polyu.hk, hongwei.yong@polyu.edu.hk,} \\
{\tt\small \{17901410r, jianqi.ma\}@connect.polyu.hk, cslzhang@comp.polyu.edu.hk}}
\maketitle


\begin{abstract}
The recently developed transformer networks have achieved impressive performance in image denoising by exploiting the self-attention (SA) in images. However, the existing methods mostly use a relatively small window to compute SA due to the quadratic complexity of it, which limits the model's ability to model long-term image information. In this paper, we propose the spatial-frequency attention network (SFANet) to enhance the network's ability in exploiting long-range dependency.
For spatial attention module (SAM), we adopt dilated SA to model long-range dependency.
In the frequency attention module (FAM), we exploit more global information by using Fast Fourier Transform (FFT) by designing a window-based frequency channel attention (WFCA) block to effectively model deep frequency features and their dependencies. To make our module applicable to images of different sizes and keep the model consistency between training and inference, we apply window-based FFT with a set of fixed window sizes. In addition, channel attention is computed on both real and imaginary parts of the Fourier spectrum, which further improves restoration performance. The proposed WFCA block can effectively model image long-range dependency with acceptable complexity.  Experiments on multiple denoising benchmarks demonstrate the leading performance of SFANet network.
\end{abstract}

\section{Introduction}
\label{sec:intro}
\begin{figure}[!t]
\centering
\vspace{0.6cm}
\begin{overpic}[width=0.48\textwidth]{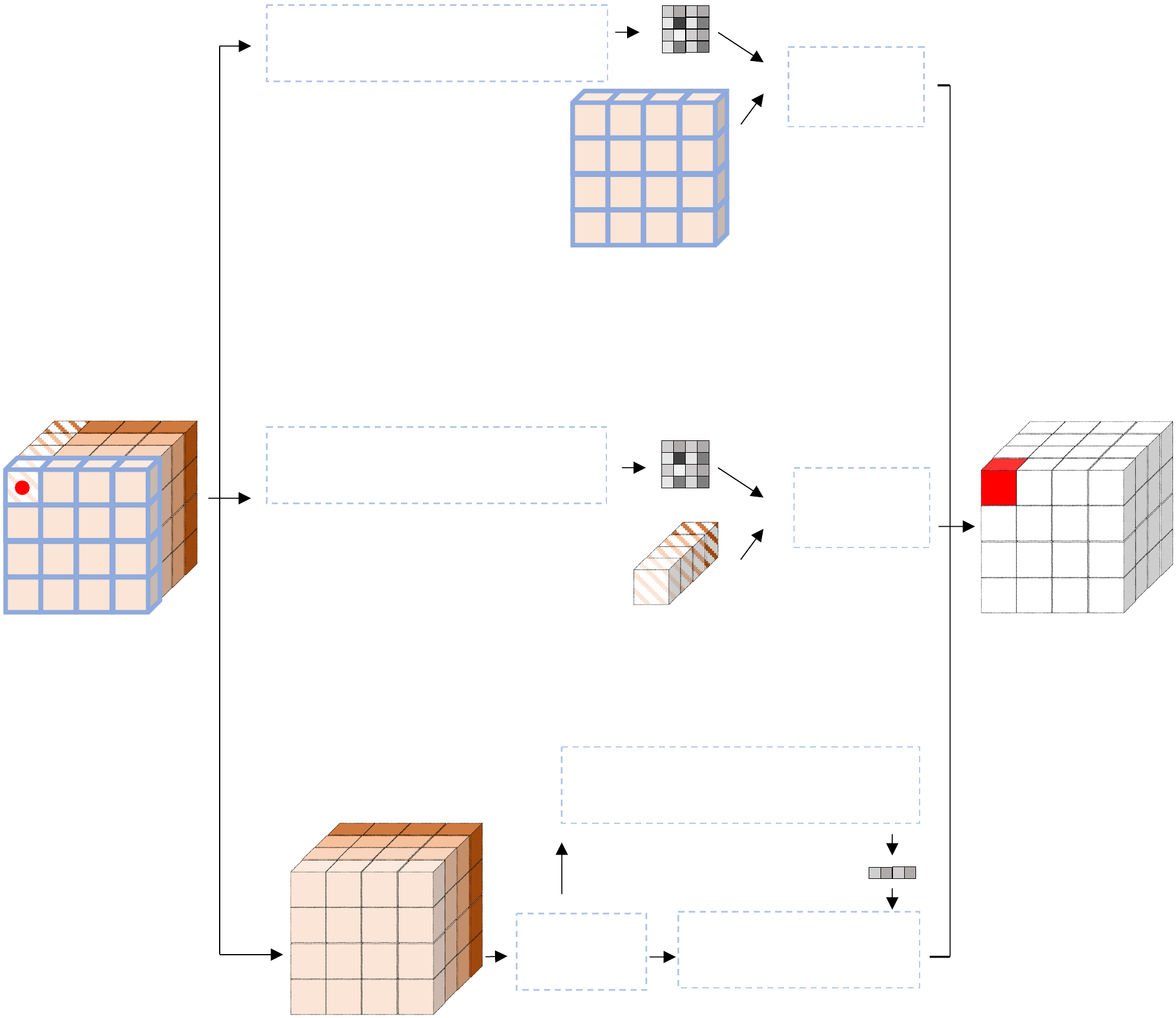}
    \put(-3,40){\color{black}{\footnotesize N}}
    \put(5,31){\color{black}{\footnotesize M}}
    \put(0,50){\color{black}{\footnotesize C}}
    \put(24.2,82){\color{black}{\scriptsize MatMul along spatial}}
    \put(28,76.5){\color{black}{\scriptsize $\boldsymbol{\mathcal{O}({(NM)}^2)}$}}
    \put(62,83.5){\color{black}{\tiny SoftMax}}
    \put(68.5,78){\color{black}{\scriptsize MatMul}}
    \put(55,87){\color{black}{\scriptsize Attention Map $\in \mathbb{ R}^{C\times NM\times NM}$}}
    \put(32, 61){\color{black}{\scriptsize (a) SA along the spatial dimension}}
    
    \put(23.4,46){\color{black}{\scriptsize MatMul along channel}}
    \put(62,47){\color{black}{\tiny SoftMax}}
    \put(68.4,42){\color{black}{\scriptsize MatMul}}
    \put(42,53){\color{black}{\scriptsize Attention Map $\in \mathbb{ R}^{C\times C\times NM}$}}
    \put(32, 31){\color{black}{\scriptsize (b) SA along the channel dimension}}
    
    \put(49.2,19){\color{black}{\scriptsize Squeeze and Excitation}}
    \put(46.5,5){\color{black}{\scriptsize Conv}}
    \put(61.5,5){\color{black}{\scriptsize Dot Product}}
    \put(54,12){\color{black}{\tiny Attention map $\in \mathbb{R}^C$}}
    \put(19,3){\color{black}{\scriptsize $\mathcal{F}$}}
    \put(22,18.5){\color{black}{\scriptsize Fourier spectrum}}
    \put(5,-3){\color{black}{\scriptsize $\boldsymbol{\mathcal{O}({(NM)\log{(NM)}})}$}}
    \put(79,2.5){\color{black}{\scriptsize $\mathcal{F}^{-1}$}}
    \put(26, -8){\color{black}{\scriptsize (c) Window-based frequency channel attention}}
\end{overpic} \vspace{0.2cm}
\caption{Illustration of different attention modules.  (a) Self-attention (SA) along the spatial dimension; (b) SA along the channel dimension and (c) our proposed window-based frequency channel attention (WFCA). For simplicity, we use deep features in an $N\times M$ window with $C$ channels as the example to illustrate different attentions. 
In WFCA, $\mathcal{F}$ and $\mathcal{F}^{-1}$ represent 2D FFT and the inverse 2D FFT, respectively. The frequency features in complex values and the channel attention (CA) is applied on both real and imaginary parts of the Fourier spectrum. }
\label{fig:compare_attention}
\vspace{-0.2cm}
\end{figure}

Image denoising is a classical yet fundamental problem in low-level vision, aiming to reconstruct a clean image from its noisy observation. The deep convolutional neural network (CNN) based denoising methods~\cite{zhang2017beyond,zhang2018ffdnet,plotz2018neural,liu2018non,guo2019toward} developed in recent years have substantially improved the denoising performance. However, the convolution (Conv) layer has a limited receptive field, making the CNN methods less effective to model image long-range dependency. Very recently, the transformer-based methods have achieved great success in image denoising~\cite{liang2021swinir,zamir2022restormer} by employing the self-attention (SA) mechanism to exploit long-range feature dependency.
SwinIR~\cite{liang2021swinir} calculates SA within a small window (see Fig.~\ref{fig:compare_attention}(a)) and achieves much better denoising performance than CNNs. However, calculating SA on small spatial windows of size $8\times 8$ or $16\times 16$ restricts the model's ability to exploit long-range information. Directly increasing the window size to compute SA is highly expensive due to quadratic complexity $\mathcal{O}({(NM)}^2)$ of SA computation for a window of height $N$ and width $M$. Restormer~\cite{zamir2022restormer} calculates SA along the channel dimension to decrease the computational complexity of SA. The structure is shown in Fig.~\ref{fig:compare_attention}(b). Restormer's attention map focuses mainly on modeling channel dependence. It has $1\times 1$ receptive field along the channel dimension in each SA layer, limiting its capacity to utilize long-range spatial information. Therefore, how to design a more effective module to model image long-range dependency deserves further investigation.

\begin{figure*}[!t]
\centering
\vspace{0.6cm}
\begin{overpic}[width=1.0\textwidth]{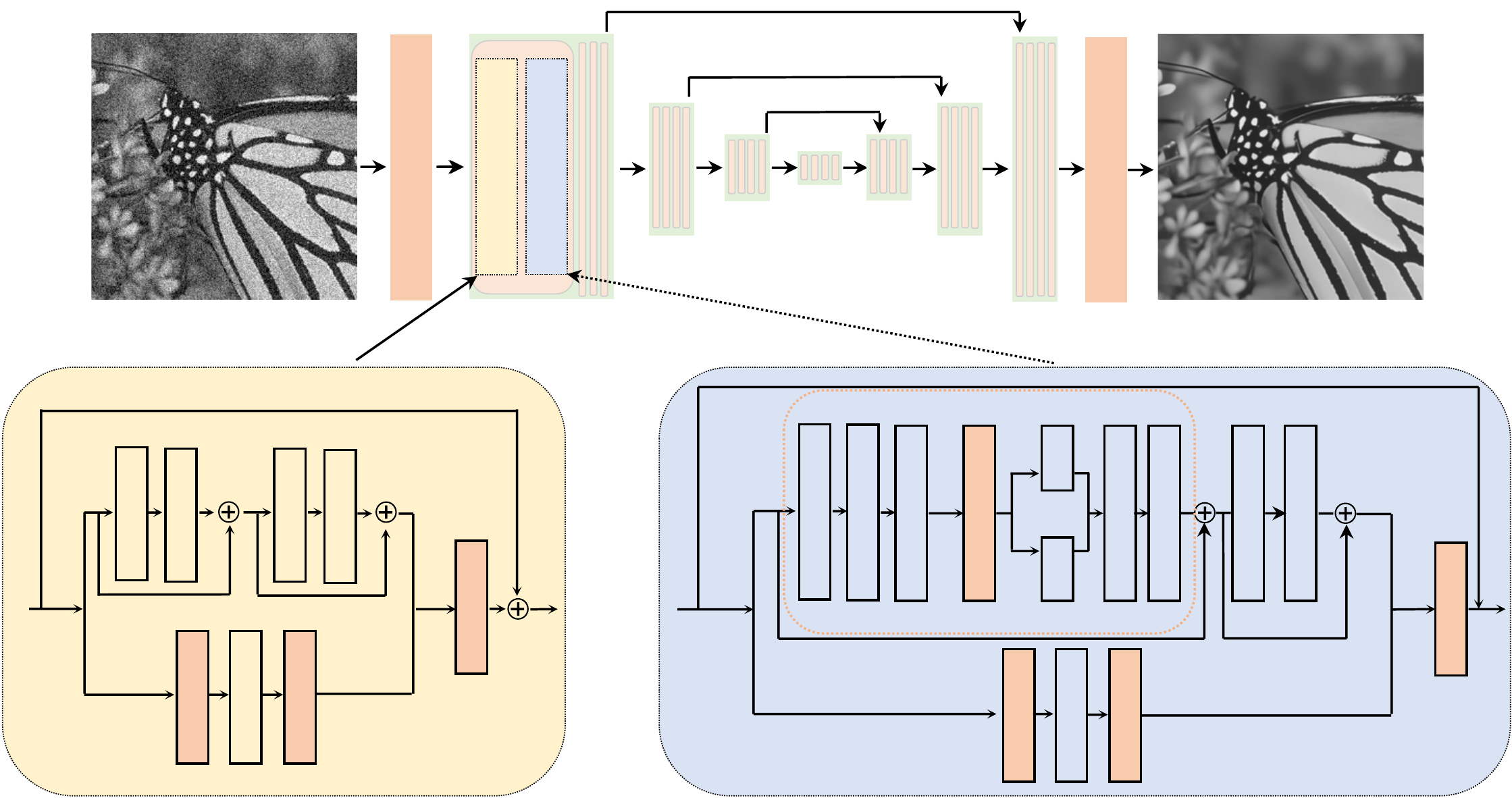}
    \put(10,31){\color{black}{\normalsize Noisy Image}}
    \put(79,31){\color{black}{\normalsize Denoised Image}}
    \put(47.2,48.5){\color{black}{\normalsize Skip Connections}}
    
    \put(26.3,39.5){\color{black}{\normalsize \rotatebox{90}{Conv}}}
    \put(72.2,39.5){\color{black}{\normalsize \rotatebox{90}{Conv}}}
    \put(31.9,39.5){\color{black}{\normalsize \rotatebox{90}{SAM}}}
    \put(35.3,39.5){\color{black}{\normalsize \rotatebox{90}{FAM}}}
    \put(5.0,-2){\color{black}{\normalsize (a) Spatial Attention Module (SAM)}}
    \put(58,-2){\color{black}{\normalsize (b) Frequency Attention Module (FAM)}}
    
    \put(3.3,13.8){\color{black}{\small \rotatebox{90}{Split}}}
    \put(28,13){\color{black}{\small \rotatebox{90}{Concat}}}
    \put(8,15){\color{black}{\footnotesize \rotatebox{90}{LayerNorm}}}
    \put(11.3,14.5){\color{black}{\small \rotatebox{90}{window SA}}}
    \put(18.5,15){\color{black}{\footnotesize \rotatebox{90}{LayerNorm}}}
    \put(21.8,16.5){\color{black}{\small \rotatebox{90}{MLP}}}
    
    \put(12,4.5){\color{black}{\small \rotatebox{90}{Conv}}}
    \put(15.5,4.5){\color{black}{\small \rotatebox{90}{ReLU}}}
    \put(19,4.5){\color{black}{\small \rotatebox{90}{Conv}}}
    \put(30.3,11){\color{black}{\small \rotatebox{90}{Conv}}}

    \put(47.5,13.5){\color{black}{\small \rotatebox{90}{Split}}}
    \put(93,13){\color{black}{\small \rotatebox{90}{Concat}}}
    \put(66.7,3.7){\color{black}{\small \rotatebox{90}{Conv}}}
    \put(70.2,3.7){\color{black}{\small \rotatebox{90}{ReLU}}}
    \put(73.6,3.7){\color{black}{\small \rotatebox{90}{Conv}}}
    \put(95.3,10.5){\color{black}{\small \rotatebox{90}{Conv}}}
    
    \put(53.2,15){\color{black}{\small \rotatebox{90}{LayerNorm}}}
    \put(56.4,14.5){\color{black}{\footnotesize \rotatebox{90}{Patch Partition}}}
    \put(53.3,11.5){\color{black}{\footnotesize \emph{window-based frequency CA (WFCA)}}}
    \put(59.5,15.7){\color{black}{\small \rotatebox{90}{2D FFT}}}
    \put(61.3,19){\color{black}{\small \rotatebox{90}{$[x_{f}^{re},x_{f}^{im}]$}}}
    \put(64.2,13.3){\color{black}{\scriptsize \rotatebox{90}{Conv+ReLU+Conv}}}
    \put(66.3,13.5){\color{black}{\small \rotatebox{90}{$x_{f1}^{im}$}}}
    \put(66.3,21.5){\color{black}{\small \rotatebox{90}{$x_{f1}^{re}$}}}
    \put(69.3,14){\color{black}{\small \rotatebox{90}{CA}}}
    \put(69.3,21.3){\color{black}{\small \rotatebox{90}{CA}}}
    \put(73.5,15.7){\color{black}{\small \rotatebox{90}{2D IFFT}}}
    \put(76.2,14.2){\color{black}{\footnotesize \rotatebox{90}{Patch Merging}}}
    \put(81.7,15.2){\color{black}{\small \rotatebox{90}{LayerNorm}}}
    \put(85.2,17){\color{black}{\small \rotatebox{90}{MLP}}}
\end{overpic}
\vspace{0.01cm}
\caption{Illustration of our SFANet. (a) illustrates the structure of spatial attention module (SAM), where we employ window-based self-attention (window SA)~\cite{liang2021swinir,zhang2022efficient,zhang2022efficient}. To increase the receptive field of SAM in shallow features, we replace window SA by using our proposed multi-scale dilated self-attention (MDSA) block on the first scale of UNet structure. The detail of MDSA is described in Fig.~\ref{figDWSA}. (b) illustrates the structure of the frequency attention module (FAM). In FAM, window-based frequency channel attention (WFCA) is designed to effectively model deep frequency feature. }
\label{figOverall}
\end{figure*}

Beyond modeling dependency in the spatial domain, we propose to exploit long-range information of the image in the frequency domain. It is well-known that each frequency component in Fourier spectrum is computed from a nearly global receptive field, and it can be efficiently calculated using the Fast Fourier Transform (FFT). The complexity of FFT is $\mathcal{O}((NM)\log{(NM)})$ for an image of size $N\times M$. Some studies~\cite{li2018frequency,zhang2022swinfir} have been reported to employ FFT in deep learning for image restoration. These methods learn several convolutional layers in frequency domain without adaptively modeling feature dependency. In addition, they apply FFT on the entire deep feature maps. However, this will introduce the mismatch problem of frequency resolution in training and inference, and hence degrade the image restoration performance. Specifically, the Fourier spectrum $x_f(u,v)$, where $u$ and $v$ respectively denote the horizontal and vertical frequency, is obtained by projecting the image feature onto the basis functions $e^{-j2\pi(\frac{un}{N} + \frac{vm}{M})}$, where $n=1,...,N$ and $m=1,...,M$ are the horizontal and vertical index. One can see that the frequency resolution of $x_f$ is determined by the input feature size $N\times M$, \ie, $\frac{2\pi}{N}$ and $\frac{2\pi}{M}$. We experimentally prove that, if the model is trained by applying FFT on the features of some size, but applied to image features of different sizes in inference, the mismatch of frequency resolution will happen and this will lead to much performance degradation. 

To tackle the above problem, we propose the spatial-frequency attention network (SFANet) for image denoising to enhance the network's ability in exploiting long-term information. SFANet consists of frequency attention module (FAM) and spatial attention module (SAM). For FAM, we propose a window-based frequency channel attention (WFCA) block to exploit long-range image dependency effectively in frequency domain. In the WFCA block, the feature is first evenly partitioned into patches with some fixed size, \eg, $N\times M$. FFT is then applied to these patches to extract frequency features. In this way, we can ensure the consistency of frequency resolution during training and inference. We then compute channel attention~\cite{hu2018squeeze,anwar2019real,zamir2020learning,guo2021joint} in frequency domain along frequency feature channels (see Fig.~\ref{fig:compare_attention} (c)). As each grid in the Fourier spectrum contains global information of spatial features, using Conv layers and CA in frequency domain can more effectively model long-range spatial dependence than Restormer, which calculates SA along channel dimensions on spatial features. In addition, previous methods~\cite{li2018frequency,zhang2022swinfir} learn simple Conv layers on real part of Fourier spectrum, which loses a part of frequency information, while we utilize both real and imaginary parts of the Fourier spectrum to model comprehensive amplitude and phase information of frequency features.
Considering that the complexity of FFT is $\mathcal{O}((NM)\log{(NM)})$, it is efficient to set the window size of WFCA to a large number, such as $N=M=64$ in our experiments, to exploit more global information than previous SA-based methods~\cite{liang2021swinir,zamir2022restormer,zhang2022efficient}. For SAM, we utilize dilated SA to model dependency in spatial domain. Experiments demonstrate that our SFANet delivers state-of-the-art performance on multiple denoising benchmarks.

\section{Related Work}
\label{related}
\subsection{Image Denoising}
The goal of image denoising is to recover a clean image from its noisy observation. The advancement of deep convolutional neural networks (CNNs) has led to great improvement on denoising performance. DnCNN~\cite{zhang2017beyond}, MemNet~\cite{tai2017memnet} and FFDNet~\cite{zhang2018ffdnet} are among the pioneer works of CNN-based denoising by adopting a plain network topology and a residual learning mechanism, which achieve significantly better results than traditional hand-crafted prior based methods. MWCNN~\cite{liu2018multi} and CBDNet~\cite{guo2019toward} used a UNet-based structure for image denoising by learning hierarchical multi-scale feature representations. Inspired by model-based image denoising techniques, some deep learning methods utilize image self-similarity priors by introducing a non-local module~\cite{liu2018non,plotz2018neural}. 

Recently, transformer models have also been introduced in image denoising tasks to exploit image long-range dependency by computing the self-attention (SA) in feature domain ~\cite{chen2021pre,liang2021swinir,zhang2022efficient,zamir2022restormer}. IPT~\cite{chen2021pre} presents a pre-trained Transformer model, which can be used for various downstream image processing tasks including denoising. Since the computational cost of SA grows quadratically with the input feature size, SwinIR~\cite{liang2021swinir} and ELAN~\cite{zhang2022efficient} calculate SA on small spatial windows of size $8\times 8$ or $16\times 16$ with a shifting mechanism. However, utilizing small window size restricts the model ability to exploit long-range information. Restormer~\cite{zamir2022restormer} calculates the SA in the channel space to reduce the computational cost but weakens the capability of structure and spatial information modeling. In this paper, we propose to utilize Fourier transformation in deep features to more effectively model long-range feature dependency for image denoising with log-linear complexity. 

\subsection{Frequency Learning in Low-level Vision}
A few works have been proposed to employ Fourier transform in deep learning for low level vision~\cite{xia2020basis,fritsche2019frequency,li2021learning,pang2020fan,chen2019drop,xie2021learning,fuoli2021fourier}. First, since convolution operation in image domain is equivalent to dot multiplication in the frequency domain, Fast Fourier Transform (FFT) can be used to accelerate the computation of Conv layers with a large kernel size~\cite{xia2020basis}. Secondly, in the frequency domain, the high-frequency components represent image textures and details, while low-frequency components represent flat and smooth areas. Therefore, some approaches~\cite{fritsche2019frequency,li2021learning,pang2020fan,chen2019drop,xie2021learning} divide images into distinct frequency intervals and employ different or dynamic network structures to handle different frequency information. 
In addition, Fourier transformation can be utilized to design loss functions to improve the image high-frequency details for better perceptual quality~\cite{fuoli2021fourier}. However, most of the existing methods only utilize FFT to assist network learning and do not directly model dependency on deep frequency features. 

Recently, deep frequency features have been directly modeled for exploiting image long-range information ~\cite{li2018frequency,zhang2022swinfir}. However, they perform FFT on the entire image, which leads to the mismatch issue when evaluating images with different sizes. Also,  only the Conv layers are utilized to extract frequency features\cite{li2018frequency,zhang2022swinfir}, which is not adaptive to the input content. To overcome these issues, we propose the window-based frequency channel attention (WFCA) block, which is more suitable to model long-range information in frequency domain.

\begin{figure*}[!t]
\centering
\vspace{0.6cm}
\begin{overpic}[width=0.5\textwidth]{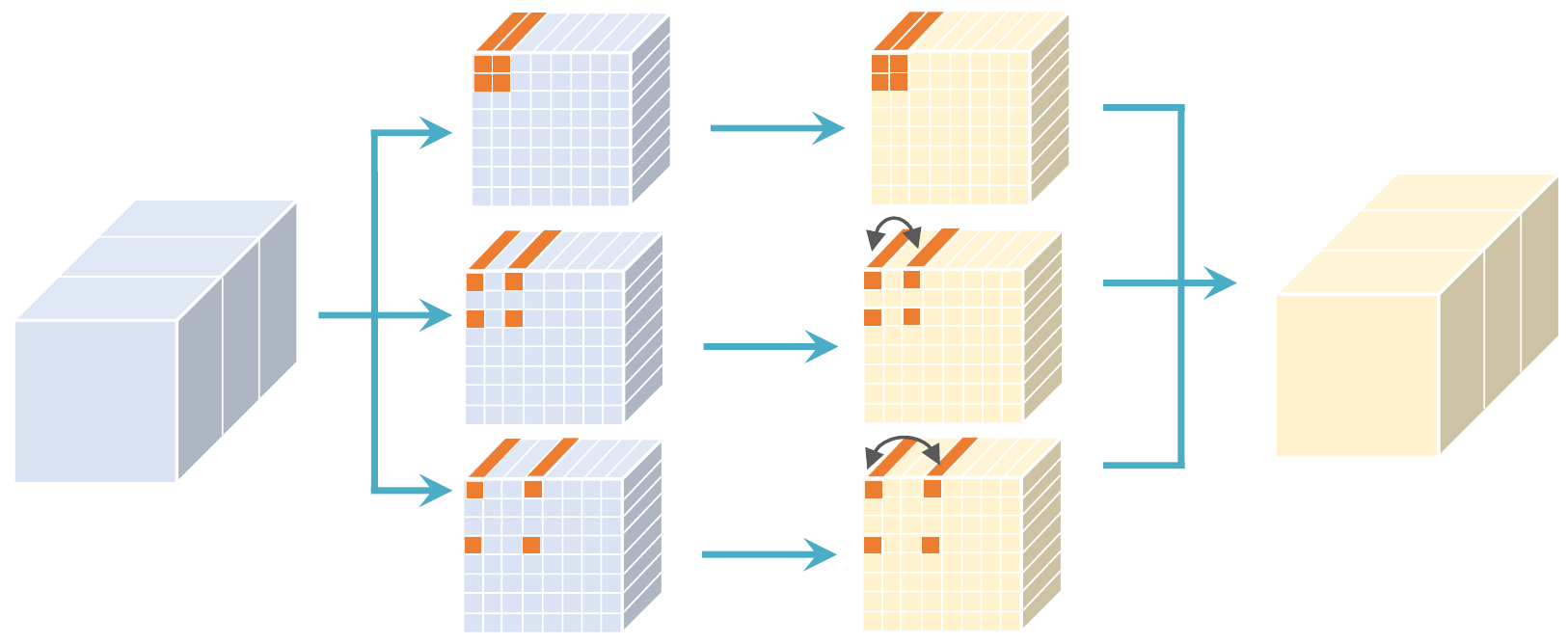}
	\put(50,27.5){\color{black}{\tiny 2 pixels}}
	\put(50,13.5){\color{black}{\tiny 4 pixels}}
	
	\put(43.3,36.5){\color{black}{\scriptsize dilated SA}}
	\put(44.5,34){\color{black}{\scriptsize ($s=1$)}}
	\put(43.0,22.5){\color{black}{\scriptsize dilated SA}}
	\put(44.5,20){\color{black}{\scriptsize ($s=2$)}}
	\put(43.0,9){\color{black}{\scriptsize dilated SA}}
	\put(44.5,6.5){\color{black}{\scriptsize ($s=4$)}}
	
	\put(40,-5){\color{black}{\footnotesize (a) MDSA}}
\end{overpic}\hspace{1cm}
%
\begin{overpic}[width=0.1\textwidth]{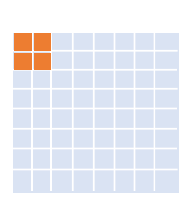}
	\put(-15,105){\color{black}{\scriptsize window size = $b$}}
	\put(-15,90){\color{black}{\scriptsize RF = $b\cdot b$}}
	\put(-0,-25){\color{black}{\footnotesize (b) window SA}}
	
\end{overpic}\hspace{0.5cm}
%
\begin{overpic}[width=0.1\textwidth]{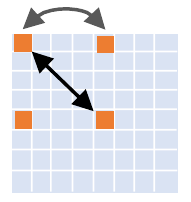}
	\put(-15,120){\color{black}{\scriptsize distance = $\frac{W}{b}$}}
	\put(-15,105){\color{black}{\scriptsize RF = nearly global}}
	\put(45,55){\color{black}{\scriptsize too far}}
	\put(8,-25){\color{black}{\footnotesize (c) grid SA}}
\end{overpic}\hspace{0.5cm}
%
\begin{overpic}[width=0.1\textwidth]{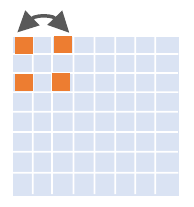}
	\put(-15,120){\color{black}{\scriptsize distance = $s$}}
	\put(-15,105){\color{black}{\scriptsize RF = $bs\cdot bs$}}
	\put(-0,-25){\color{black}{\footnotesize (d) dilated SA}}
\end{overpic}\hspace{0.5cm}

\vspace{0.3cm}
\caption{Illustration of Multi-scale Dilated Self-attention (MDSA). RF represents the receptive field.}
\label{figDWSA}
\vspace{-0.2cm}
\end{figure*}
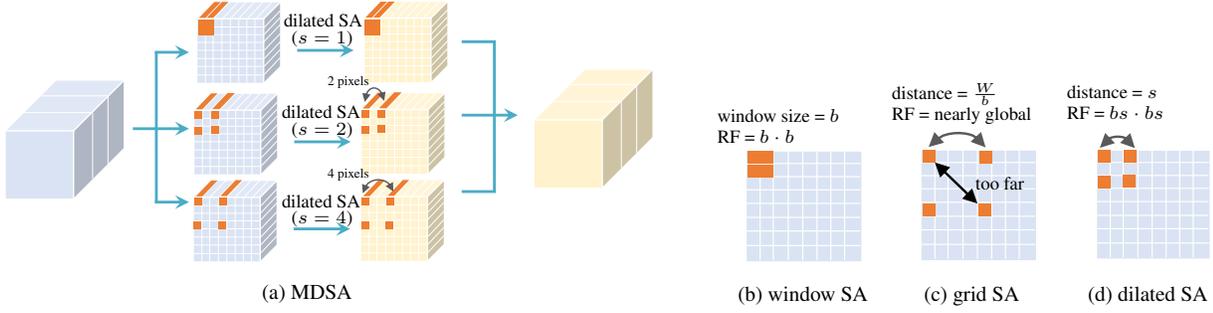

\section{Method}
\label{method}
\subsection{Overall Network Structure}
Fig.~\ref{figOverall} illustrates our SFANet. For the input noise image $y$, we first obtain a shallow feature $x_s$ by using one $3\times 3$ convolution layer, $x_s = F_{s}(y)$. Then we investigate deep features using a UNet structure, denoted as $x_d = F_u(x_s)$. The UNet structure $F_u$ consists of four scales with symmetric skip connections. Strided and transpose convolutions are employed as downsampling and upsampling operators in UNet, respectively. $F_u$ is composed of two main modules, \ie, spatial attention module (SAM) and frequency attention module (FAM), which utilize spatial attention and frequency attention respectively to explore long-range image dependency. Then, the clean image is reconstructed as $\hat{x} = F_r(x_d)$, where $F_r$ is the reconstruction module and we simply use one $3\times 3$ convolution layer as $F_r$.

\textbf{SAM:} Fig.~\ref{figOverall}(a) shows the structure of SAM. The input feature $x$ is evenly separated into two branches, denoted as $x_a$ and $x_c$, which are processed by window-based SA (window SA)~\cite{liang2021swinir,zhang2022efficient} and Conv layers to model deep feature simultaneously. The SA branch can be formulated as:
\begin{align}
    x_a &= \text{WSA}(\text{LN}(x_a)) + x_a, \nonumber \\
    x_a &= F(\text{LN}(x_a)) + x_a, 
\end{align}
where $\text{WSA}(\cdot)$ is window SA, $\text{LN}(\cdot)$ represents layer normalization and $F(\cdot)$ contains two $1\times 1$ Conv layers with ReLU non-linearity. Following \cite{liang2021swinir}, window SA utilizes the multi-head strategy with window size $8\times 8$. For the Conv branch, the output is obtained using: 
\begin{equation}
    x_c = F_2(\delta(F_1(x_c))),
\label{eq:conv_branch}
\end{equation}
where $F_1 \in \mathbb{R}^{C\times rC}$ and $F_2\in \mathbb{R}^{rC\times C}$ are two $1\times 1$ convolution layers, $\delta$ is the ReLU function. We set $r=4$ in our experiment. Finally, the output $y$ is obtained by merging $x_a$ and $x_c$ via a $1\times 1$ Conv layer.

To expand the receptive field of window SA in shallow features without involving additional computations, we design a multi-scale dilated self-attention (MDSA) block.
In the first scale of UNet, the window SA is replaced by MDSA, which utilizes our proposed dilated self-attention (dilated SA). The structure of MDSA and dilated SA are shown in Fig.~\ref{figDWSA} and will be described in Sec.~\ref{sec:dsa}.

\textbf{FAM:} The structure of FAM is shown in Fig.~\ref{figOverall} (b). The input feature of FAM, denoted as $x$, is also firstly split into two branches, $x_{fr}$ and $x_{c}$, which are then fed to frequency and Conv branches, separately. For the frequency branch, we formulate the process as:
\begin{align}
    x_{fr} &= \text{WFCA}(x_{fr}) + x_{fr}, \nonumber \\
    x_{fr} &= F(\text{LN}(x_{fr})) + x_{fr}, 
\end{align}
where $F(\cdot)$ consists of Conv layers with ReLU function, $\text{WFCA}(\cdot)$ is our proposed window-based frequency channel attention module (WFCA). In WFCA, we design a simple but effective window-based strategy to solve the mismatch problem for inputs of varying sizes and use channel attention in the frequency domain to improve the restoration performance. By modeling deep frequency feature in WFCA, we can investigate long-range dependency with acceptable complexity. More details of WFCA will be described in Sec.~\ref{sec:wfca}. For the Conv branch, the output feature is obtained using $x_c = F_2(\delta(F_1(x_c)))$, which is in the same structure as Equ.~\ref{eq:conv_branch}. One Conv layer is used to merge $x_{fr}$ and $x_c$ and obtain the output of FAM.

\begin{table*}[!tbp]
\setlength{\abovecaptionskip}{0.2cm}
\setlength{\belowcaptionskip}{0.0cm}
\footnotesize
\caption{Performance comparison of different denoising methods on three benchmarks.} 
\label{table:awgng}
\begin{tabular}{{p{1.8cm}|p{0.8cm}|p{1.1cm} p{1.1cm} p{1.1cm} | p{1.1cm} p{1.1cm} p{1.1cm} | p{1.1cm} p{1.1cm} p{0.9cm}}}
\Xhline{0.5pt}
\centering
Method & &\multicolumn{3}{c}{Set12}  &\multicolumn{3}{c}{BSD68} &\multicolumn{3}{c}{Urban100} \\
\Xhline{0.5pt}
Noise Level & &$\sigma=15$ &$\sigma=25$ &$\sigma=50$ &$\sigma=15$ &$\sigma=25$ &$\sigma=50$ &$\sigma=15$ &$\sigma=25$ &$\sigma=50$ \\
\Xhline{1pt}
\multirow{2}{*}{DnCNN~\cite{zhang2017beyond}} &PSNR &32.85 &30.43 &27.17 &31.74 &29.23 &26.24 &32.64 &29.95 &26.26 \\
&SSIM &0.9025 &0.8617 &0.7828 &0.8907 &0.8279 &0.7189 &0.9241 &0.8781 &0.7856 \\
\Xhline{0.5pt}
\multirow{2}{*}{FFDNet~\cite{zhang2018ffdnet}} &PSNR &32.74 &30.42 &27.30 &31.64 &29.19 &26.29 &32.40 &29.90 &26.50\\
&SSIM &0.9024 &0.8631 &0.7899 &0.8902 &0.8288 &0.7239 &0.9265 &0.8879 &0.8047\\
\Xhline{0.5pt}
\multirow{2}{*}{IRCNN~\cite{zhang2017learning}} &PSNR &32.76 &30.37 &27.12 &31.64 &29.15 &26.19 &32.46 &29.80 &26.22\\
&SSIM &0.9006 &0.8598 &0.7804 &0.8882 &0.8248 &0.7169 &0.9236 &0.8831 &0.7918\\
\Xhline{0.5pt}
\multirow{2}{*}{N3Net~\cite{plotz2018neural}} &PSNR &\hspace{0.16cm} --- &30.55 &27.43 &\hspace{0.16cm} --- &29.30 &26.39 &\hspace{0.16cm} --- &30.19 &26.26 \\
&SSIM &\hspace{0.16cm} --- &\hspace{0.16cm} --- &\hspace{0.16cm} --- &\hspace{0.16cm} --- &\hspace{0.16cm} --- &\hspace{0.16cm} --- &\hspace{0.16cm} --- &\hspace{0.16cm} --- &\hspace{0.16cm} --- \\
\Xhline{0.5pt}
\multirow{2}{*}{NLRN~\cite{liu2018non}} &PSNR &33.16 &30.80 &27.64 &31.88 &29.41 &26.47 &33.45 &30.94 &27.49 \\
&SSIM &0.9070 &0.8689 &0.7980 &0.8932 &0.8331 &0.7298 &0.9354 &0.9018 &0.8279 \\
\Xhline{0.5pt}
\multirow{2}{*}{FOCNet~\cite{jia2019focnet}} &PSNR &33.07 &30.73 &27.68 &31.83 &29.38 &26.50 &33.15 &30.64 &27.40 \\
&SSIM &\hspace{0.16cm} --- &\hspace{0.16cm} --- &\hspace{0.16cm} --- &\hspace{0.16cm} --- &\hspace{0.16cm} --- &\hspace{0.16cm} --- &\hspace{0.16cm} --- &\hspace{0.16cm} --- &\hspace{0.16cm} --- \\
\Xhline{0.5pt}
\multirow{2}{*}{GCDN~\cite{valsesia2020deep}} &PSNR &33.14 &30.78 &27.60 &31.83 &29.35 &26.38 &33.47 &30.95 &27.41  \\
&SSIM &0.9072 &0.8687 &0.7957 &0.8933 &0.8332 &0.7389 &0.9358 &0.9020 &0.8160 \\
\Xhline{0.5pt}
\multirow{2}{*}{DAGL~\cite{mou2021dynamic}} &PSNR &33.28 &30.93 &27.81 &31.93 &29.46 &26.51 &33.79 &31.39 &27.97 \\
&SSIM &0.9100 &0.8720 &0.8042 &0.8953 &0.8366 &0.7334 &0.9393 &0.9093 &0.8423 \\
\Xhline{0.5pt}
\multirow{2}{*}{DRUNet~\cite{zhang2021plug}} &PSNR &32.25 &30.94 &27.90 &31.91 &29.48 &26.59 &33.44 &31.11 &27.96\\
&SSIM &0.9098 &0.8732 &0.8096 &0.8952 &0.8371 &0.7378 &0.9376 &0.9082 &0.8483\\
\Xhline{0.5pt}
\multirow{2}{*}{SwinIR~\cite{liang2021swinir}} &PSNR &33.36 &31.01 &27.91 &31.97 &29.50 &26.58 &33.70 &31.30 &27.98\\
&SSIM &0.9110 &0.8741 &0.8096 &0.8960 &0.8376 &0.7377 &0.9391 &0.9094 &0.8474\\
\Xhline{0.5pt}
\multirow{2}{*}{Restormer~\cite{zamir2022restormer}} &PSNR &33.42 &31.08 &28.00 &31.96 &29.52 &26.62 &33.79 &31.46 &28.29\\
&SSIM &0.9127 &0.8759 &0.8121 &0.8964 &\textbf{0.8388} &0.7398 &0.9401 &0.9121 &0.8554 \\
\Xhline{0.5pt}
\multirow{2}{*}{Ours} &PSNR &\textbf{33.46} &\textbf{31.10} &\textbf{28.05} &\textbf{32.01} &\textbf{29.55} &\textbf{26.68} &\textbf{33.97} &\textbf{31.67} &\textbf{28.70} \\
&SSIM &\textbf{0.9131} &\textbf{0.8761} &\textbf{0.8136} &\textbf{0.8968} &0.8386 &\textbf{0.7416} &\textbf{0.9414} &\textbf{0.9146} &\textbf{0.8638}\\

\Xhline{1pt}
\end{tabular} 
\vspace{-0.2cm}
\end{table*}

\subsection{Window-based Frequency Channel Attention}
\label{sec:wfca}
The architecture of the WFCA block is shown in Fig~\ref{figOverall} (b). The input feature with size $B\times C\times H\times W$, is denoted as $x$, which is normalized by layer normalization. To overcome the mismatch problem of previous methods~\cite{li2018frequency,zhang2022swinfir}, we first segment $x$ uniformly into non-overlapped patches $p_n$ with window size $N$. The size of $p_n$ is $(B\frac{H}{N}\frac{W}{N})\times C\times N\times N$. FFT is then used to transform the deep feature into the frequency domain:
{\small \begin{align}
    x_f(b,c,u,v) &= \mathcal{F}(p_n) \nonumber\\
        &= \sum_{h=0}^{N-1} \sum_{w=0}^{N-1} p_n(b,c,h,w)e^{-j2\pi(\frac{uh}{N} + \frac{vw}{N})},
\end{align}} 

\noindent where $\mathcal{F}(\cdot)$ is the Fast Fourier Transform, $b$ and $c$ are indices of batch and channel dimensions, $u$ and $v$ denote $u$-th horizontal and $v$-th vertical spatial frequencies in the Fourier spectrum $x_f$, which is in complex value and can be expressed as $x_f = x_{f}^{re} + x_{f}^{im}\cdot i$ with $x_{f}^{re}$ and $x_{f}^{im}$ being the real and imaginary parts. One can see that the frequency resolution of $x_{f}$ is $\frac{2\pi}{N}$, which is irrelevant to image size.

To extract deep frequency features, previous methods~\cite{li2018frequency,zhang2022swinfir} ignore the propriety of complex number and only apply Conv layer on real part of $x_f$. Referring to \cite{hu2020dccrn,zhao2022frcrn}, the output of the complex convolution $x_{f1}$ can be expressed as:
\begin{align}
    x_{f1}^{re} &= (x_{f}^{re}\otimes W^{re}) - (x_{f}^{im}\otimes W^{im}), \nonumber \\
    x_{f1}^{im} &= (x_{f}^{re}\otimes W^{im}) + (x_{f}^{im}\otimes W^{re}), 
\label{eq:compleConv}
\end{align}
where $W^{re}$ and $W^{im}$ are the real and imaginary parts of complex convolutional kernels. One can see that the real and imaginary parts of $x_{f1}$ are affected by both the real and imaginary parts of $x_{f}$. To simplify the training process, we rewrite Equ.~\ref{eq:compleConv} into the following equation:
\begin{equation}
    x_{f1}^{re}, x_{f1}^{im} = F([x_{f}^{re}, x_{f}^{im}]), 
\end{equation}
in which $[,]$ is the concatenation operator and $F(\cdot)$ is the $1\times 1$ Conv layer. Then, we utilize channel attention (CA)~\cite{hu2018squeeze,anwar2019real,zamir2020learning,guo2021joint} in the Fourier spectrum to adaptively model deep frequency features. We firstly obtain two $C\times 1\times 1$ channel descriptors for real and imaginary parts by performing global average pooling on $x_{f1}^{re}$ and $x_{f1}^{im}$ independently. These channel descriptors are denoted as $z^{re}$ and $z^{im}$. Then we obtain the channel attention map $z_{a}^{re}$ and $z_{a}^{im}$ by using two Conv layers with the sigmoid function. The enhanced frequency feature is obtained by rescaling $x_{f1}$ with the attention map:
\begin{equation}
    x_{f2} = (x_{f1}^{re}\cdot(1 + z^{re}_{a})) + (x_{f1}^{im}\cdot(1 + z^{im}_{a}))\cdot i.
\end{equation}
Then $x_{f2}$ is transformed back to image domain by using $p_m' = \mathcal{F}^{-1}(x_{f2})$, where $\mathcal{F}^{-1}(\cdot)$ is the inverse FFT. The output of WFCA, denoted as $f'$, is to rearrange back $p_m'$ into $B\times C\times H\times W$. 

\begin{figure*}[!t]
\setlength{\abovecaptionskip}{0.1cm}
\setlength{\belowcaptionskip}{-0.cm}
\begin{subfigure}{0.245\linewidth}
  \centering
  \begin{subfigure}{1\linewidth}
    \includegraphics[width=1\textwidth]{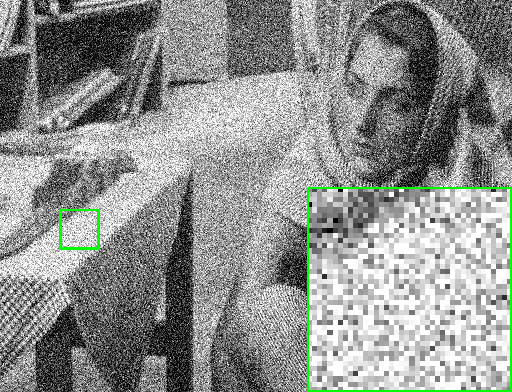}
  \end{subfigure} 
\caption{Noisy Image}
\end{subfigure} 
\begin{subfigure}{0.245\linewidth}
  \centering
  \begin{subfigure}{1\linewidth}
    \includegraphics[width=1\textwidth]{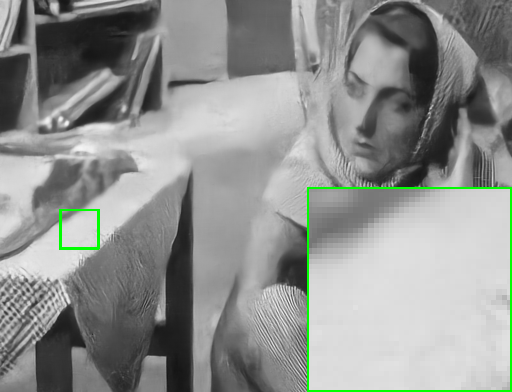}
  \end{subfigure} 
\caption{DnCNN}
\end{subfigure} 
\begin{subfigure}{0.245\linewidth}
  \centering
  \begin{subfigure}{1\linewidth}
    \includegraphics[width=1\textwidth]{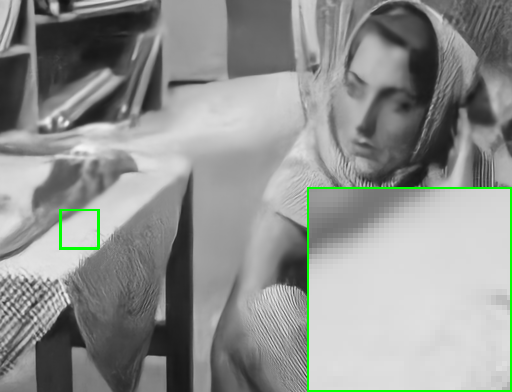}
  \end{subfigure} 
\caption{FFDNet}
\end{subfigure}
\begin{subfigure}{0.245\linewidth}
  \centering
  \begin{subfigure}{1\linewidth}
    \includegraphics[width=1\textwidth]{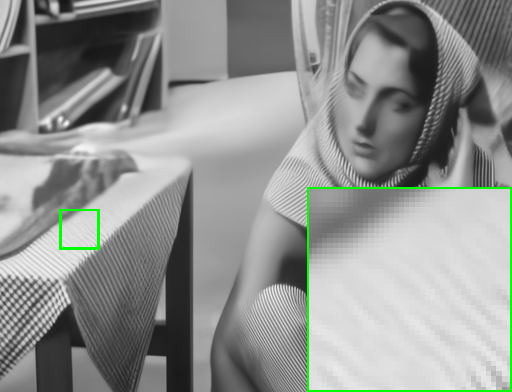}
  \end{subfigure} 
\caption{DRUNet}
\end{subfigure}

\begin{subfigure}{0.245\linewidth}
  \centering
  \begin{subfigure}{1\linewidth}
    \includegraphics[width=1\textwidth]{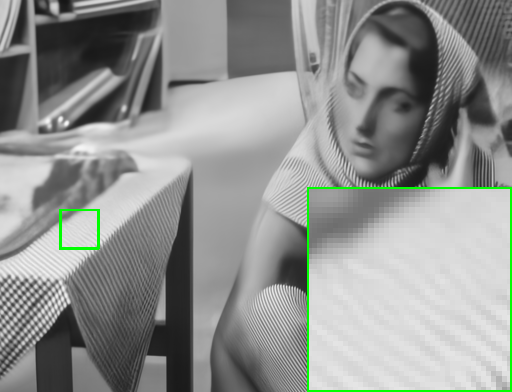}
  \end{subfigure} 
\caption{SwinIR}
\end{subfigure} 
\begin{subfigure}{0.245\linewidth}
  \centering
  \begin{subfigure}{1\linewidth}
    \includegraphics[width=1\textwidth]{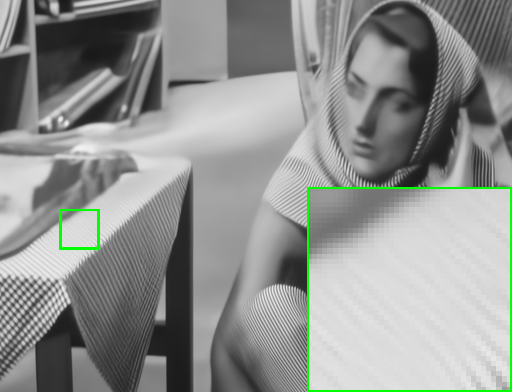}
  \end{subfigure} 
\caption{Restormer}
\end{subfigure} 
\begin{subfigure}{0.245\linewidth}
  \centering
  \begin{subfigure}{1\linewidth}
    \includegraphics[width=1\textwidth]{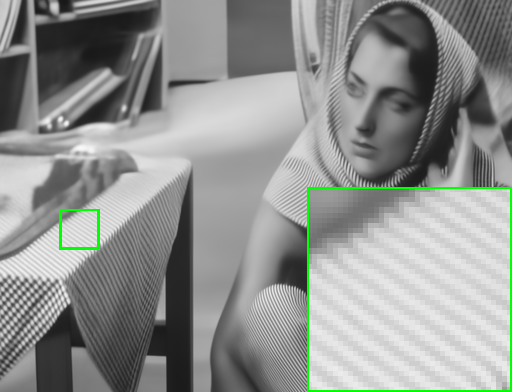}
  \end{subfigure} 
\caption{Ours}
\end{subfigure} 
\begin{subfigure}{0.245\linewidth}
  \centering
  \begin{subfigure}{1\linewidth}
    \includegraphics[width=1\textwidth]{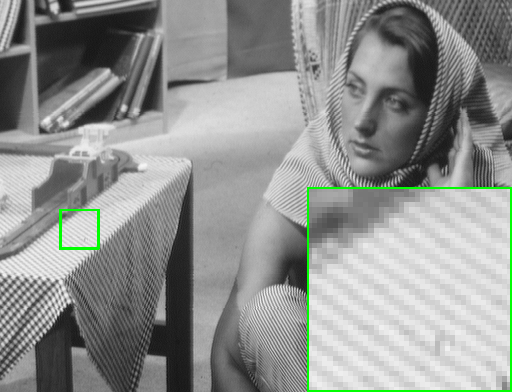}
  \end{subfigure} 
\caption{Ground-truth}
\end{subfigure} 

\caption{Denoising results on image "\emph{Barbara}" from Set12 dataset with noise level 50 by different methods.}
\label{fig:set12}
\vspace{-0.2cm}
\end{figure*}

\subsection{Dilated Self-attention}
\label{sec:dsa}
To enlarge the receptive field of window SA, some approaches~\cite{zhang2022efficient,chen2022activating} directly enlarge the window size from $8\times 8$ to $16 \times 16$ to employ more input pixels for reconstruction, which quadruples the computational cost. Recently, \cite{zhao2021improved,tu2022maxim} propose a variant of window SA, \ie, multi-axis self-attention (MASA), to efficiently activate long-range pixels. The MASA initially converts a feature with size $(H, W, C)$ to $(b\times b, \frac{H}{b}\times \frac{W}{b}, C)$ and calculates SA along the 1st and 2nd axes, where $b$ is the window-size. For the axis-1, SA is calculated on the local window area $b\times b$, which is equivalent to the regular window SA (see Fig.~\ref{figDWSA} (b)). Calculating SA along the axis-2 is called grid SA, which is shown in Fig.~\ref{figDWSA} (c). The grid SA is calculated on pixels with stride $\frac{H}{b}$ and $\frac{W}{b}$ along vertical and horizontal directions. Though grid SA can model long-range spatial dependency with acceptable complexity, we discover that grid SA is not ideal for low-level vision. First, features used for image restoration normally have high resolution, which makes the stride $\frac{H}{b}$ quite large and decreases the dependency between pixels. More importantly, the stride of grid SA, \ie, $\frac{H}{b}$, is related to image size, which also causes the mismatch issue when the input images have various sizes. 

To solve these problems, we design dilated SA block and show its structure in Fig.~\ref{figDWSA} (d). The input tensor of size $(H\times W\times C)$ is firstly rearranged into a tensor of shape $(s\times s, b\times b, \frac{H}{b\cdot s}\times \frac{W}{b\cdot s}, C)$, where $b$ and $s$ represent window size and stride value, respectively. Then we calculate SA along the second axis, \ie $b\times b$. As shown in Fig.~\ref{figDWSA} (d), dilated SA models spatial dependency on $b\times b$ pixels with stride $s$. Compared with window SA with window size $b$, the receptive field of dilated SA enlarges to $bs\times bs$ without introducing additional calculations. For $s=1$, the dilated attention is equivalent to window SA. For the MDSA block, we utilize dilated SA with different strides. As shown in Fig.~\ref{figDWSA} (a), the input feature tensor is separated into three sub-features with size $(H\times W\times C/3)$. Then three sub-features are processed by the dilated SA with stride $s=1, 2, 4$, separately. Finally, the processed sub-features are concatenated and merged by one $1\times 1$ Conv layer.

\begin{figure*}
\setlength{\abovecaptionskip}{0.1cm}
\setlength{\belowcaptionskip}{-0.cm}

\begin{subfigure}{0.245\linewidth}
    \includegraphics[width=1\textwidth]{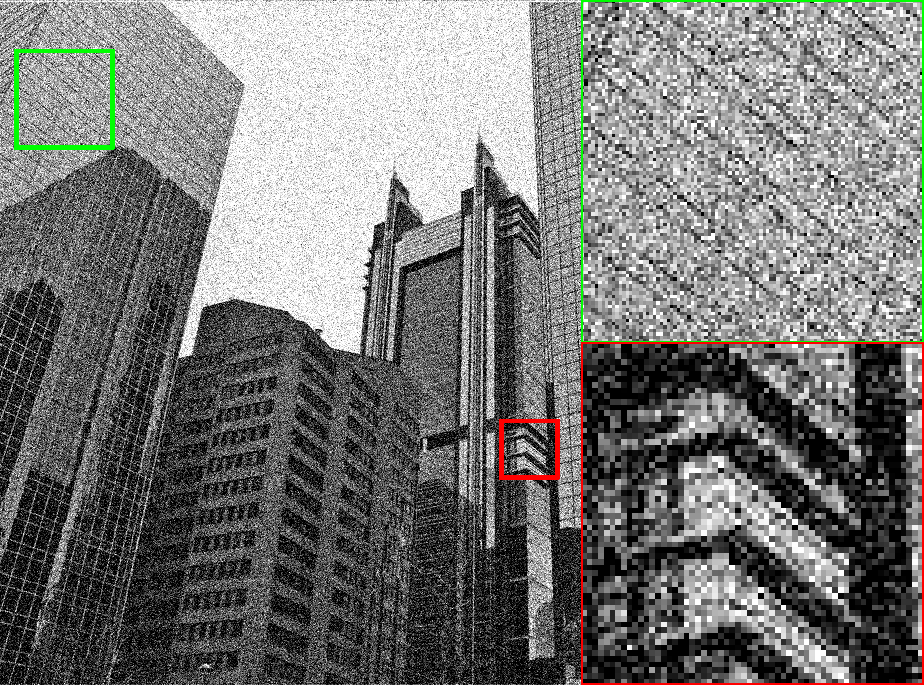}
\caption{Noisy image}
\end{subfigure}
\begin{subfigure}{0.245\linewidth}
    \includegraphics[width=1\textwidth]{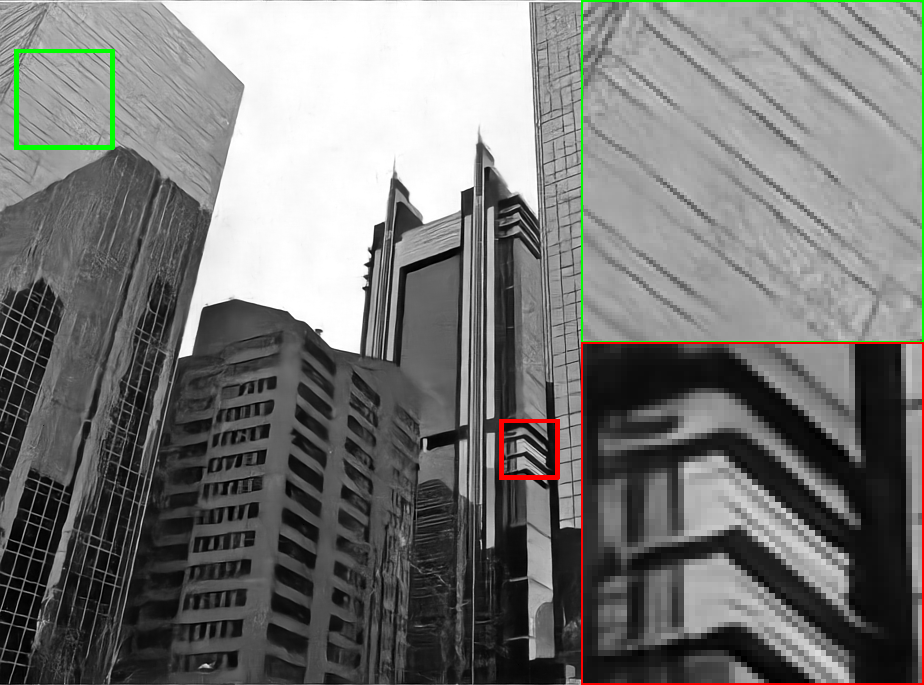}
\caption{DnCNN}
\end{subfigure}
\begin{subfigure}{0.245\linewidth}
    \includegraphics[width=1\textwidth]{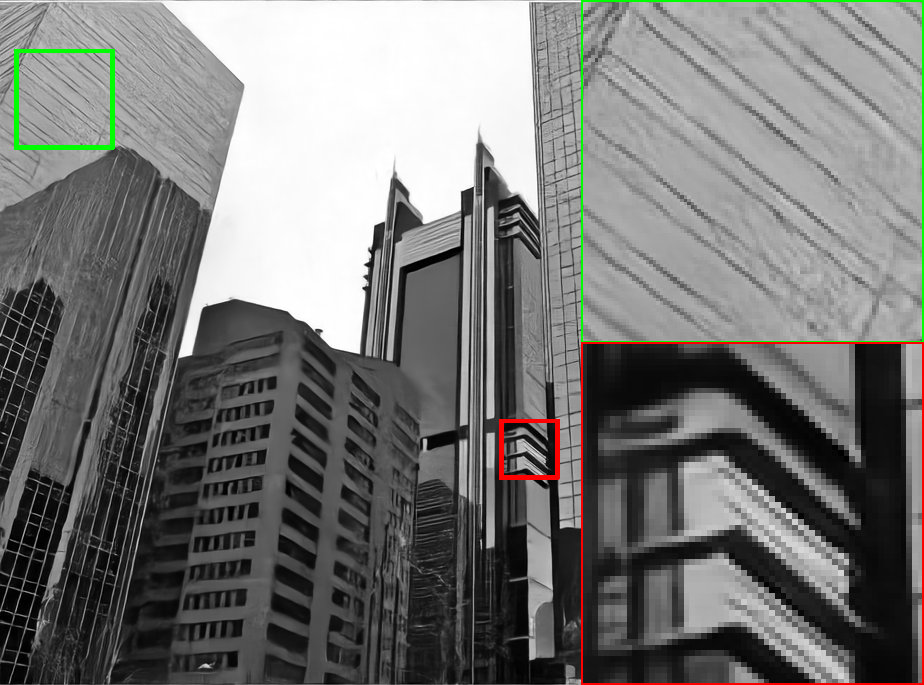}
\caption{FFDNet}
\end{subfigure} 
\begin{subfigure}{0.245\linewidth}
    \includegraphics[width=1\textwidth]{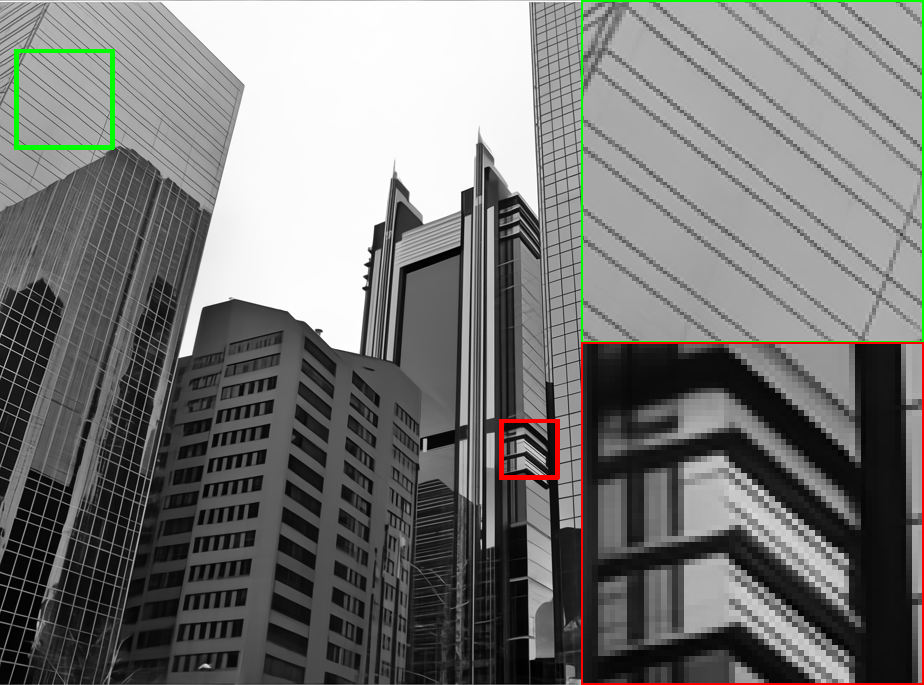}
\caption{DRUNet}
\end{subfigure} 

\begin{subfigure}{0.245\linewidth}
    \includegraphics[width=1\textwidth]{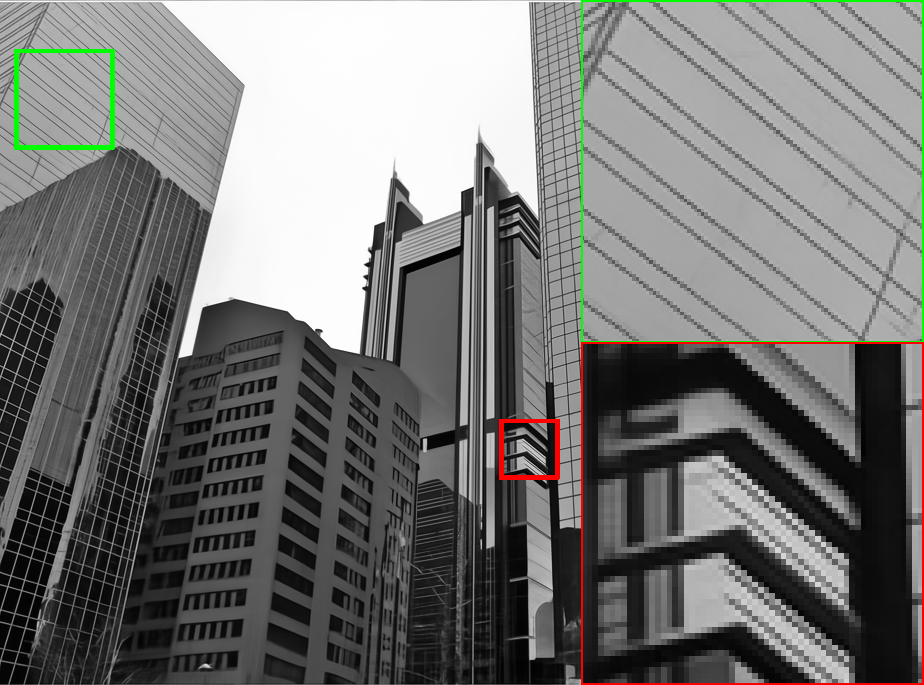}
\caption{SwinIR}
\end{subfigure}
\begin{subfigure}{0.245\linewidth}
    \includegraphics[width=1\textwidth]{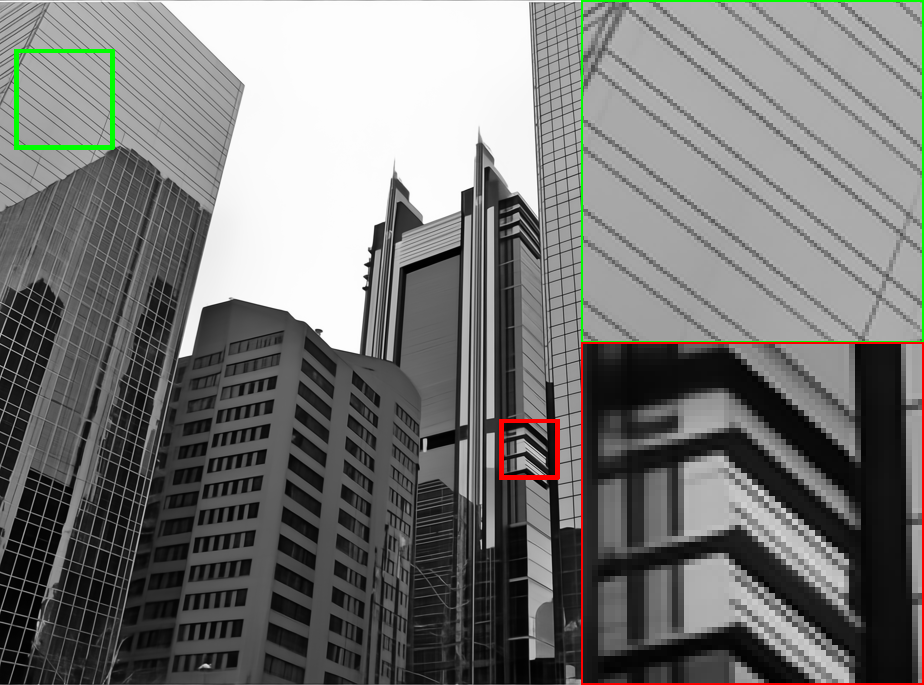}
\caption{Restormer}
\end{subfigure}
\begin{subfigure}{0.245\linewidth}
    \includegraphics[width=1\textwidth]{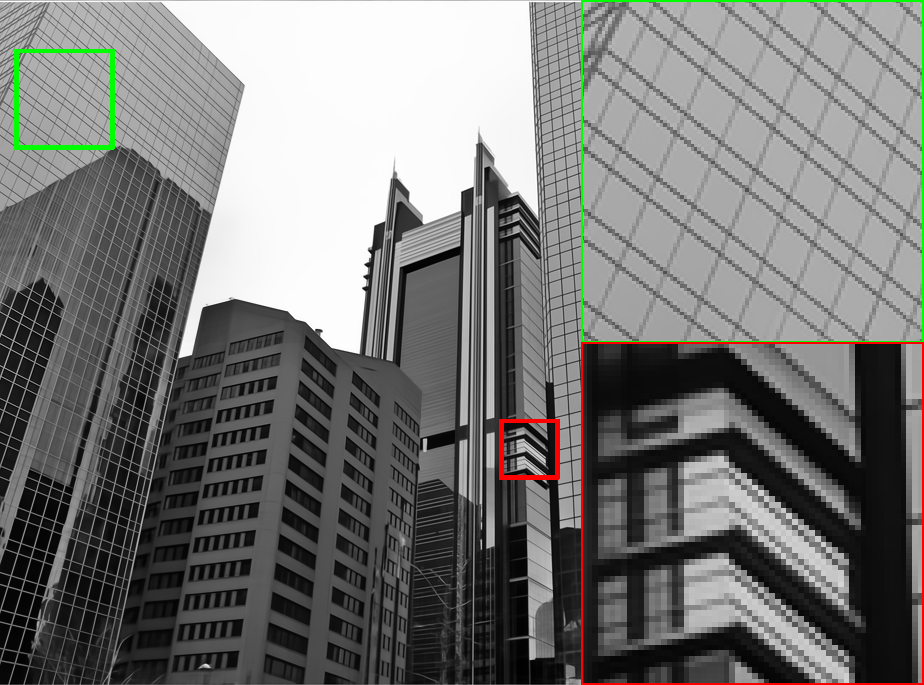}
\caption{SFANet (Ours)}
\end{subfigure} 
\begin{subfigure}{0.245\linewidth}
    \includegraphics[width=1\textwidth]{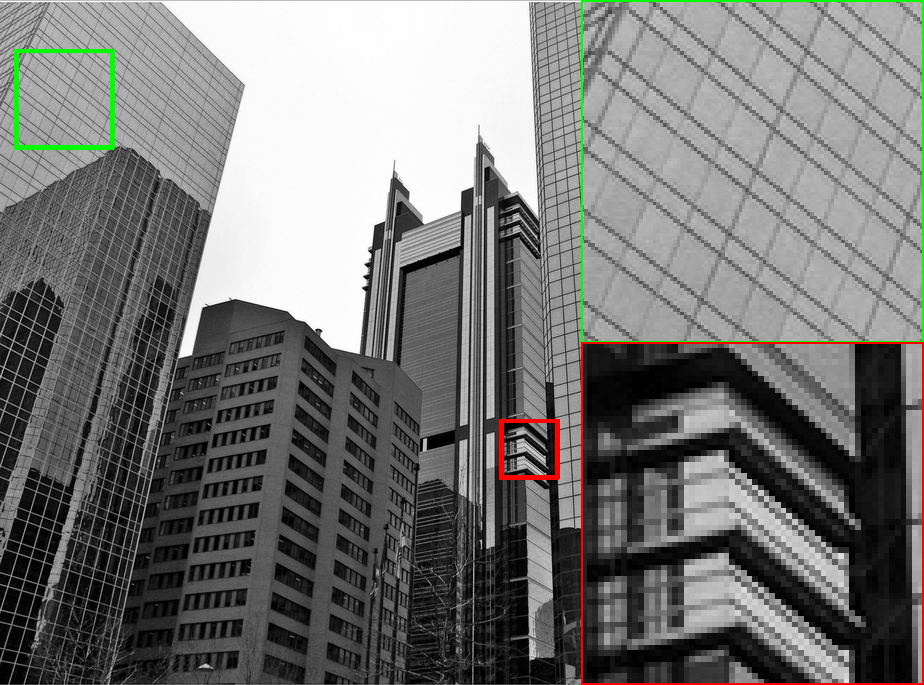}
\caption{Ground-truth}
\end{subfigure} 

\caption{Denoising results on Urban100 with noise level 50 by different methods.}
\label{fig:urban100}
\end{figure*}

\section{Experiments}
\subsection{Experiments setting} The channel numbers of UNet are set as 64, 128, 256 and 512 for the four scales. We utilize different window size $N = 64, 32, 16, 8$ for WFCA blocks on different UNet scales. The Charbonnier penalty function is used as the loss function, and Adam optimizer is used. The learning rate is initialized as $1\times 10^{-4}$, and it is decays by 0.5 for every 200,000 iterations and ends to $3.125\times 10^{-6}$. 

The training dataset consists of Waterloo Exploration Dataset~\cite{ma2016waterloo}, DIV2K~\cite{agustsson2017ntire}, Flick2K~\cite{lim2017enhanced}, BSD400~\cite{martin2001database} and OST~\cite{wang2018sftgan}. The training patch size is $192\times 192$. The network for noise level 25 is firstly trained, and the models for noise levels 15 and 50 are initialized by using the pre-trained model with noise level 25 and then trained with half learning iterations. To evaluate our method, we compare our method with the state-of-the-art methods: DnCNN~\cite{zhang2017beyond}, FFDNet~\cite{zhang2018ffdnet}, IRCNN~\cite{zhang2017learning}, N3Net~\cite{plotz2018neural}, NLRN~\cite{liu2018non}, FOCNet~\cite{jia2019focnet}, GCDN~\cite{valsesia2020deep}, DAGL~\cite{mou2021dynamic}, DRUNet~\cite{zhang2021plug}, SwinIR~\cite{liang2021swinir}, and Restormer~\cite{zamir2022restormer}. 

\subsection{Results on AWGN Denoising}
Table~\ref{table:awgng} shows the quantitative comparison on Set12~\cite{zhang2017beyond}, BSD68~\cite{martin2001database} and Urban100~\cite{huang2015single}. One can see that methods that use nonlocal module (\ie, NLRN, GCDN and DAGL), UNet structure (\ie, FOCNet, DRUNet and Restormer) or SA module (\ie, SwinIR and Restormer) to model long-range information achieve substantial improvement over methods with pure CNN structure (\ie, DnCNN, FFDNet and IRCNN) on Urban100 dataset, where images have rich repetitive structures. By using WFCA to adaptively model deep frequency features, our method obtains average $\sim$0.3dB PSNR improvement over the state-of-the-art Restormer on Urban100 for all noise levels, which proves the effectiveness of WFCA in modeling long-range information for image denoising.

The qualitative comparisons on Set12 and Urban100 are presented in Figs.~\ref{fig:set12} and \ref{fig:urban100}, respectively. By using the SA module, SwinIR can recover more structures than DnCNN and FFDNet. Restormer further utilizes SA in the UNet structure and achieves better performance. Restormer calculates SA along channel dimension, which lowers the ability of modeling spatial dependency and leads to worse performance on repetitive pattern recovery than our method. By increasing the receptive field to $64\times 64$ in one WFCA module and using the channel attention to adaptively model deep frequency features, our SFANet can recover clearer and richer structures than comparison methods. 

\subsection{Results on Texture Denoising}
To further prove the effectiveness of our SFANet on texture restoration, we evaluate methods on several wildly used texture image datasets, \ie Describable Textures Dataset (DTD)~\cite{cimpoi2014describing}, UIUC~\cite{lazebnik2005sparse} and Kyberge~\cite{Kylberg2011c}. 
Table~\ref{table:texture_results} and Fig.~\ref{fig:texture_com} show the quantitative and qualitative comparisons, respectively. One can see that our SFANet obtains the best PSNR/SSIM measures. By using WFCA to model long-range dependency, SFANet can recover much clearer the repetitive structures over other methods.

\begin{table}[!t]
\setlength{\abovecaptionskip}{0.2cm}
\setlength{\belowcaptionskip}{0.0cm}
\small
\centering
\caption{Quantitative comparison of different methods on texture datasets with noise level 50.}
\label{table:texture_results}
\begin{tabular}{l | c c c}
\hline
&DTD &UIUC &Kyberge\\
\hline
DnCNN &25.76/0.7170 &24.16/0.6515 &25.00/0.7579 \\
FFDNet &25.99/0.7284 &24.27/0.6583 &25.26/0.7681 \\
DRUNet &26.82/0.7571 &24.69/0.6886 &26.10/0.7926 \\
SwinIR &26.84/0.7573 &24.70/0.6883 &26.23/0.7969 \\
Restormer &27.16/0.9658 &24.80/0.6964 &26.75/0.8119 \\
\hline
SFANet &\textbf{27.35/0.7691} &\textbf{24.89/0.7021} &\textbf{26.94/0.8213} \\
\hline
\end{tabular} 
\end{table}

\begin{figure*}[!t]
\setlength{\abovecaptionskip}{0.0cm}
\setlength{\belowcaptionskip}{-0.cm}
\begin{subfigure}{0.162\linewidth}
  \centering
  \begin{subfigure}{1\linewidth}
    \includegraphics[width=1\textwidth]{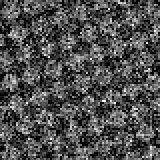}
  \end{subfigure} 
\end{subfigure} 
\begin{subfigure}{0.162\linewidth}
  \centering
  \begin{subfigure}{1\linewidth}
    \includegraphics[width=1\textwidth]{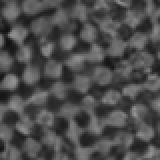}
  \end{subfigure} 
\end{subfigure} 
\begin{subfigure}{0.162\linewidth}
  \centering
  \begin{subfigure}{1\linewidth}
    \includegraphics[width=1\textwidth]{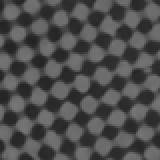}
  \end{subfigure} 
\end{subfigure} 
\begin{subfigure}{0.162\linewidth}
  \centering
  \begin{subfigure}{1\linewidth}
    \includegraphics[width=1\textwidth]{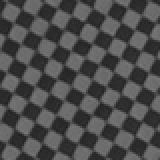}
  \end{subfigure} 
\end{subfigure} 
\begin{subfigure}{0.162\linewidth}
  \centering
  \begin{subfigure}{1\linewidth}
    \includegraphics[width=1\textwidth]{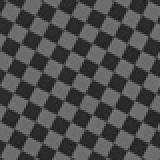}
  \end{subfigure} 
\end{subfigure} 
\begin{subfigure}{0.162\linewidth}
  \centering
  \begin{subfigure}{1\linewidth}
    \includegraphics[width=1\textwidth]{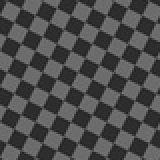}
  \end{subfigure} 
\end{subfigure}

\begin{subfigure}{0.162\linewidth}
  \centering
  \begin{subfigure}{1\linewidth}
    \includegraphics[width=1\textwidth]{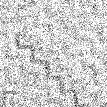}
    \caption{Noisy Image}
  \end{subfigure} 
\end{subfigure} 
\begin{subfigure}{0.162\linewidth}
  \centering
  \begin{subfigure}{1\linewidth}
    \includegraphics[width=1\textwidth]{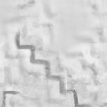}
    \caption{DnCNN}
  \end{subfigure} 
\end{subfigure} 
\begin{subfigure}{0.162\linewidth}
  \centering
  \begin{subfigure}{1\linewidth}
    \includegraphics[width=1\textwidth]{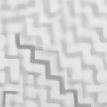}
    \caption{SwinIR}
  \end{subfigure} 
\end{subfigure} 
\begin{subfigure}{0.162\linewidth}
  \centering
  \begin{subfigure}{1\linewidth}
    \includegraphics[width=1\textwidth]{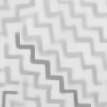}
    \caption{Restormer}
  \end{subfigure} 
\end{subfigure} 
\begin{subfigure}{0.162\linewidth}
  \centering
  \begin{subfigure}{1\linewidth}
    \includegraphics[width=1\textwidth]{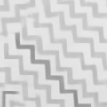}
    \caption{Ours}
  \end{subfigure} 
\end{subfigure} 
\begin{subfigure}{0.162\linewidth}
  \centering
  \begin{subfigure}{1\linewidth}
    \includegraphics[width=1\textwidth]{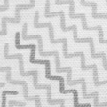}
    \caption{Ground-truth}
  \end{subfigure} 
\end{subfigure}

\caption{Visual comparisons on texture datasets with noise level 50 by different methods.}
\label{fig:texture_com}
\end{figure*}

\section{Ablation Study}
\begin{figure*}[!t]
\setlength{\abovecaptionskip}{0.0cm}
\setlength{\belowcaptionskip}{-0.cm}
\begin{subfigure}{0.138\linewidth}
  \centering
  \begin{subfigure}{1\linewidth}
    \includegraphics[width=1\textwidth]{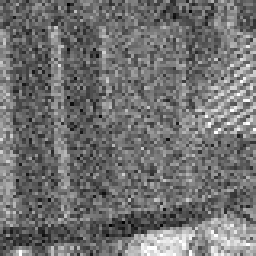}
  \end{subfigure} 
\end{subfigure} 
\begin{subfigure}{0.138\linewidth}
  \centering
  \begin{subfigure}{1\linewidth}
    \includegraphics[width=1\textwidth]{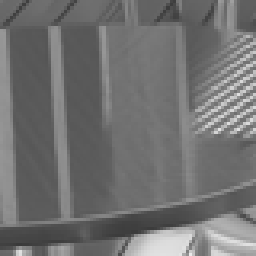}
  \end{subfigure} 
\end{subfigure} 
\begin{subfigure}{0.138\linewidth}
  \centering
  \begin{subfigure}{1\linewidth}
    \includegraphics[width=1\textwidth]{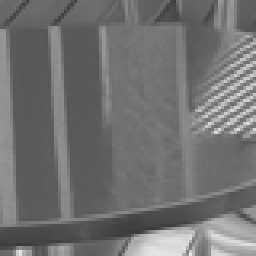}
  \end{subfigure} 
\end{subfigure} 
\begin{subfigure}{0.138\linewidth}
  \centering
  \begin{subfigure}{1\linewidth}
    \includegraphics[width=1\textwidth]{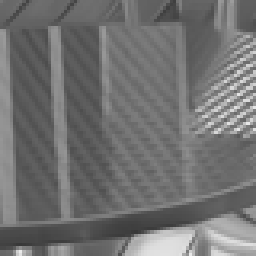}
  \end{subfigure} 
\end{subfigure} 
\begin{subfigure}{0.138\linewidth}
  \centering
  \begin{subfigure}{1\linewidth}
    \includegraphics[width=1\textwidth]{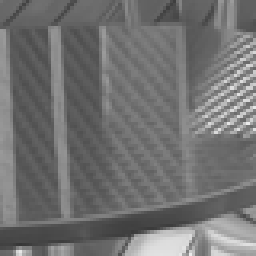}
  \end{subfigure} 
\end{subfigure} 
\begin{subfigure}{0.138\linewidth}
  \centering
  \begin{subfigure}{1\linewidth}
    \includegraphics[width=1\textwidth]{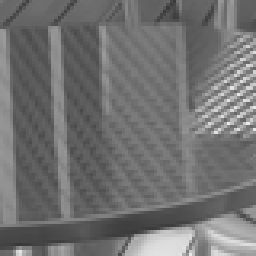}
  \end{subfigure} 
\end{subfigure}
\begin{subfigure}{0.138\linewidth}
  \centering
  \begin{subfigure}{1\linewidth}
    \includegraphics[width=1\textwidth]{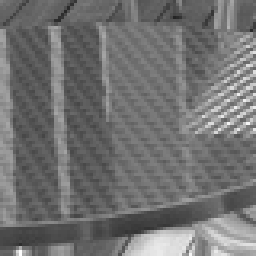}
  \end{subfigure} 
\end{subfigure} 

\begin{subfigure}{0.139\linewidth}
  \centering
  \begin{subfigure}{1\linewidth}
    \includegraphics[width=1\textwidth]{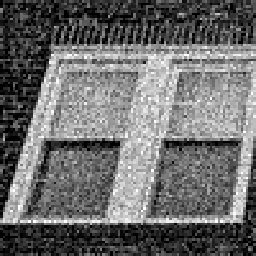}
    \caption*{Noisy Image}
  \end{subfigure} 
\end{subfigure} 
\begin{subfigure}{0.139\linewidth}
  \centering
  \begin{subfigure}{1\linewidth}
    \includegraphics[width=1\textwidth]{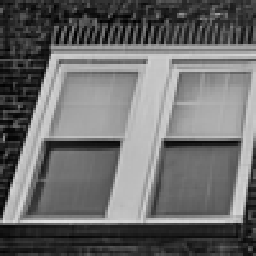}
    \caption*{SFANet(w/o WFCA)}
  \end{subfigure} 
\end{subfigure} 
\begin{subfigure}{0.139\linewidth}
  \centering
  \begin{subfigure}{1\linewidth}
    \includegraphics[width=1\textwidth]{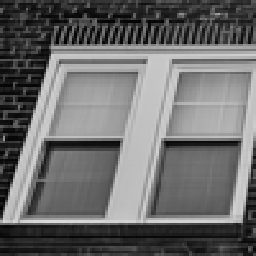}
    \caption*{SFANet(w/o W)}
  \end{subfigure} 
\end{subfigure} 
\begin{subfigure}{0.139\linewidth}
  \centering
  \begin{subfigure}{1\linewidth}
    \includegraphics[width=1\textwidth]{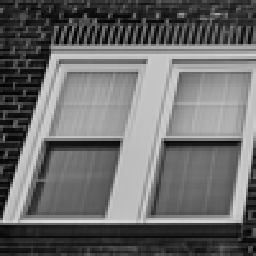}
    \caption*{SFANet(w/o CA)}
  \end{subfigure} 
\end{subfigure} 
\begin{subfigure}{0.139\linewidth}
  \centering
  \begin{subfigure}{1\linewidth}
    \includegraphics[width=1\textwidth]{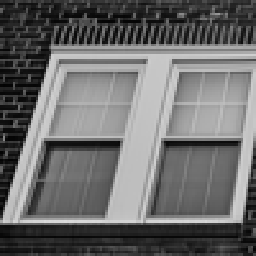}
    \caption*{SFANet(w/o IM)}
  \end{subfigure} 
\end{subfigure} 
\begin{subfigure}{0.139\linewidth}
  \centering
  \begin{subfigure}{1\linewidth}
    \includegraphics[width=1\textwidth]{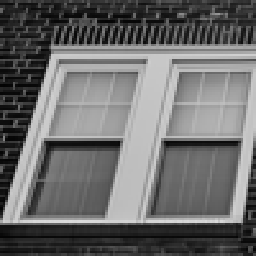}
    \caption*{SFANet}
  \end{subfigure} 
\end{subfigure}
\begin{subfigure}{0.139\linewidth}
  \centering
  \begin{subfigure}{1\linewidth}
    \includegraphics[width=1\textwidth]{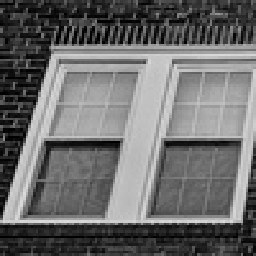}
    \caption*{Ground-truth}
  \end{subfigure} 
\end{subfigure} 

\caption{Visual comparison by using different varients of SFANet. Image is from Urban100 with noise level 25.}
\label{fig:ablation_com}
\vspace{-0.2cm}
\end{figure*}

In ablation experiments, the evaluation is performed on Set12 and Urban100 datasets with noise level 25. Table~\ref{table:ablation} shows the quantitative comparisons. The visual comparison of different variants of SFANet is shown in Fig.~\ref{fig:ablation_com}. Then we describe the effect of each component separately. 

\textbf{WFCA block.} To evaluate the role of the proposed WFCA block, we remove the frequency module in FAM by replacing WFCA with one Conv layer and denote the model as SFANet(w/o WFCA). One can see that our model can achieve $\sim$0.2dB improvement over SFANet(w/o WFCA), which proves the effectiveness of WFCA block. By modeling deep frequency features using our WFCA block, our SFANet can reconstruct more textures (see Fig.~\ref{fig:ablation_com}).

\textbf{Window-based strategy.} 
In order to overcome the mismatch problem in the frequency domain, we propose to use the window-based strategy in WFCA. To evaluate the effectiveness of the window-based strategy, we train a variant, namely SFANet(w/o W), which performs FFT on the entire deep feature and utilizes channel attention on such frequency feature. As discussed in Sec.\ref{sec:intro}, such a setting would cause the mismatch problem when inferencing images with varying size. We can see that compared with SFANet(full), the mismatch problem in SFANet(w/o W) leads to 0.06dB and 0.22dB performance degradation on Set12 and Urban100 dataset, respectively. Since the image size in urban100 (\eg, $1024\times 1024$) is very different from the training patch size (\ie, $192\times 192$), the mismatch problem is more obvious on urban100 dataset. SFANet(w/o W) obtains even worse results than SFANet(w/o WFCA) on Urban100 dataset.

From the visual comparison in Fig.\ref{fig:ablation_com}, we can also see that SFANet(w/o W) cannot recover more textures compared with SFANet(w/o WFCA). Similar experimental phenomena can also be found in \cite{zhang2022swinfir}, \ie, the frequency module is not effective when the mismatch problem exists. To overcome the mismatch problem, our SFANet(full) utilizes the window-based strategy and can obtain significantly clearer results than SFANet(w/o WFCA) with the same number of network parameters.

\textbf{Channel attention in frequency domain.} To evaluate the role of the channel attention (CA) module in our model, we train two variants, \ie, without using CA module (SFANet(w/o CA)), and without performing CA on the imaginary part of frequency spectrum (SFANet(w/o IM)). We can see that by using CA on both real and imaginary parts of the frequency spectrum, SFANet(full) can obtain 0.07dB and 0.03dB improvement on Urban100 dataset than SFANet(w/o CA) and SFANet(w/o IM), respectively. 

In Fig.~\ref{fig:ablation_com}, we can see that by using CA only on real part of the frequency spectrum, SFANet(w/o IM) can obtain denoising results with clearer structures and fewer artifacts than SFANet(w/o CA). By using both real and imaginary parts of the frequency spectrum, SFANet(full) can model comprehensive amplitude and phase information and obtain clearer denoising results. It is worth noting that, even without using CA in the frequency domain, SFANet(w/o CA) can also recover more textures than SFANet(w/o WFCA), which proves that modeling deep frequency feature with simple Conv layers can also improve the network ability in texture recovering. 

\vspace{-1mm}
\begin{table}[!t]
\setlength{\abovecaptionskip}{0.2cm}
\setlength{\belowcaptionskip}{-0.0cm}
\small
\centering
\caption{Quantitative comparison of different varients of SFANet on Set12 and Urban100 with noise level 25.}
\label{table:ablation}
\begin{tabular}{l | c c}
\hline
&Set12 &Urban100\\
\hline
SFANet(w/o WFCA) &31.02/0.8744 &31.48/0.9117 \\
SFANet(w/o W) &31.04/0.8748 &31.45/0.9116 \\
SFANet(w/o CA) &31.07/0.8756 &31.60/0.9140 \\
SFANet(w/o IM) &31.09/0.8759 &31.64/0.9145 \\

\hline
SFANet(full) &\textbf{31.10/0.8761} &\textbf{31.67/0.9146} \\
\hline
\end{tabular} 
\vspace{-0.5cm}
\end{table} 

\section{Conclusion}
\vspace{-0.2cm}
We presented a spatial-frequency attention network (SFANet) for high-performance image denoising. 
In particular, we proposed a window-based frequency channel attention (WFCA) module to effectively model image long-range dependency. WFCA utilized channel attention in the deep frequency feature domain. Since each frequency component contains nearly global information, WFCA can model more global dependency than traditional SA-based block with log-linear complexity. By using the window-based strategy, our WFCA solved the frequency resolution mismatch problem of previous methods and obtained $\sim$0.2dB improvement on Urban100. The channel attention mechanism was applied on both real and imaginary parts of the frequency spectrum to adaptively model comprehensive amplitude and phase information of deep frequency features. Dilated SA module was used in our SFANet to model long-term dependency in spatial domain. Our SFANet demonstrated clear advantages over existing methods in terms of PSNR/SSIM measures as well as visual quality in multiple image denoising benchmarks, especially on repetitive texture structures.

{\small
\bibliographystyle{ieee_fullname}
\bibliography{egbib}
}

\end{document}